\documentclass[runningheads]{llncs}
\usepackage[colorlinks=true,citecolor=blue,linkcolor=blue,urlcolor=blue]{hyperref}

\usepackage[T1]{fontenc}
\usepackage{graphicx,verbatim}
\usepackage{amsmath}
\usepackage{float}
\usepackage{amssymb}
\usepackage{multirow}
\usepackage{caption}
\usepackage{booktabs}
\usepackage{graphicx}
\usepackage{makecell}
\usepackage{pifont} 
\begin{document}
\title{CRISP: Rank-Guided Iterative Squeezing for Robust Medical Image Segmentation under Domain Shift}
\titlerunning{CRISP: Rank-Based Segmentation via Iterative Squeezing}
%
\author{Yizhou Fang\inst{1} \and
Pujin Cheng\inst{1,2} \and
Yixiang Liu\inst{1} \and
Xiaoying Tang\inst{1,3}\thanks{Corresponding authors: tangxy@sustech.edu.cn, zhoulx@sustech.edu.cn} \and
Longxi Zhou\inst{4}\textsuperscript{*} }

\authorrunning{Y. Fang et al.}

\institute{Department of Electronic and Electrical Engineering, Southern University of Science and Technology, Shenzhen, China \and
The University of Hong Kong, Hong Kong, China \and
Jiaxing Research Institute, Southern University of Science and Technology, Jiaxing, China \and
Department of Biomedical Engineering, Southern University of Science and Technology, Shenzhen, China}

\maketitle         
\begin{abstract}
Distribution shift in medical imaging remains a central bottleneck for the clinical translation of medical AI. Failure to address it can lead to severe performance degradation in unseen environments and exacerbate health inequities. Existing methods for domain adaptation are inherently limited by exhausting predefined possibilities through simulated shifts or pseudo-supervision. Such strategies struggle in the open-ended and unpredictable real world, where distribution shifts are effectively infinite. To address this challenge, we introduce an empirical law called ``Rank Stability of Positive Regions'', which states that the relative rank of predicted probabilities for positive voxels remains stable under distribution shift. Guided by this principle, we propose CRISP, a parameter-free and model-agnostic framework requiring no target-domain information. CRISP is the first framework to make segmentation based on rank rather than probabilities. CRISP simulates model behavior under distribution shift via latent feature perturbation, where voxel probability rankings exhibit two stable patterns: regions that consistently retain high probabilities (destined positives according to the principle) and those that remain low-probability (can be safely classified as negatives). Based on these patterns, we construct high-precision (HP) and high-recall (HR) priors and recursively refine them under perturbation. We then design an iterative training framework, making HP and HR progressively ``squeeze'' to the final segmentation. Extensive evaluations on multi-center cardiac MRI and CT-based lung vessel segmentation demonstrate CRISP's superior robustness, significantly outperforming state-of-the-art methods with striking HD95 reductions of up to 0.14 (7.0\% improvement), 1.90 (13.1\% improvement), and 8.39 (38.9\% improvement) pixels across multi-center, demographic, and modality shifts, respectively.

\end{abstract}
\section{Introduction}

\begin{figure*}[t] 
\centering 
\includegraphics[width=1.0\textwidth]{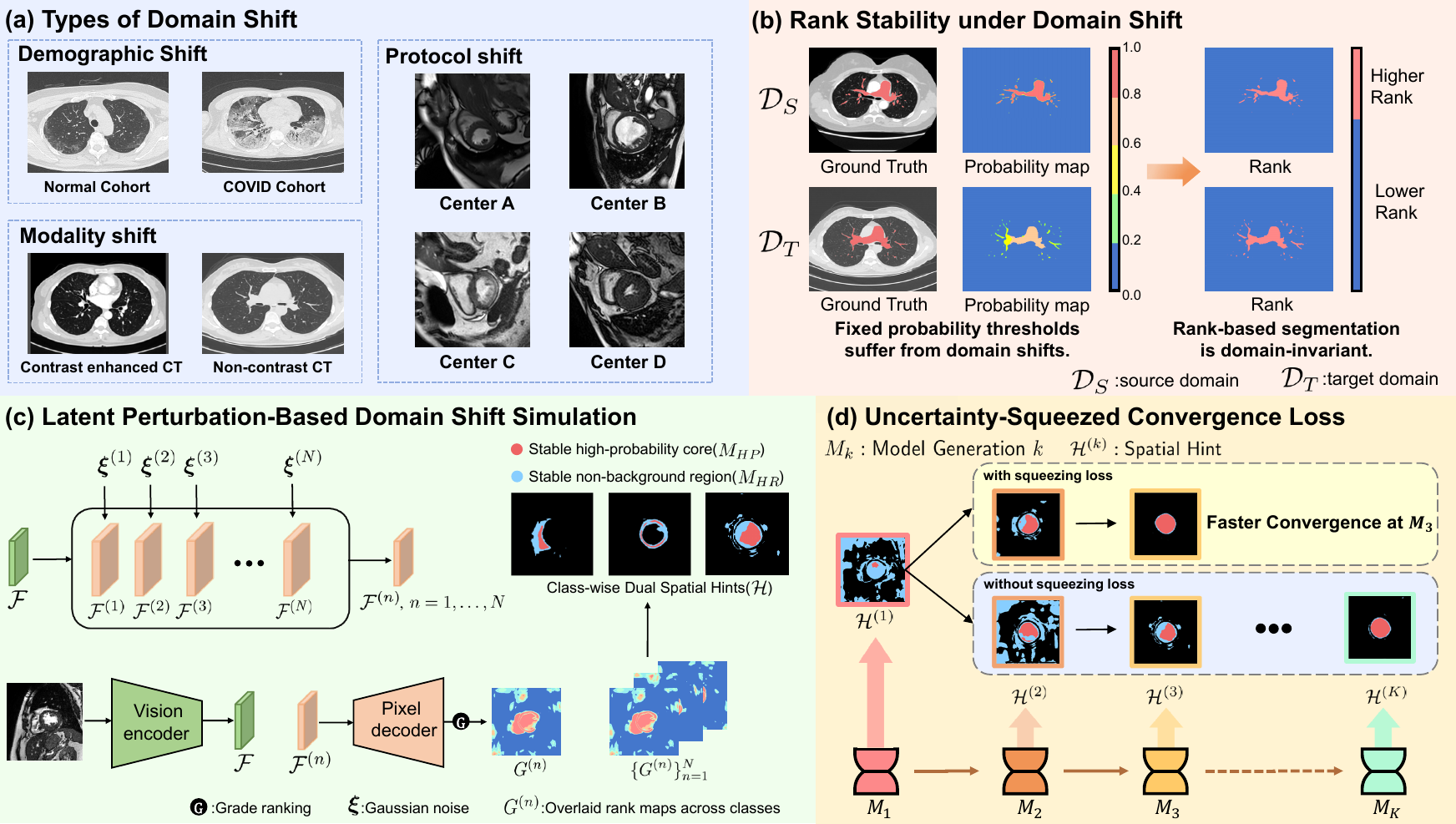} 
\caption{Overview of CRISP, a rank-based segmentation framework rather than probability-driven prediction. (a) Three representative clinical shifts: demographic, modality, and protocol variations. (b) Empirical Rank Stability: the relative rank of foreground regions remains stable despite probability drifts across domains. (c) Latent Feature Perturbation: injecting Gaussian noise into bottleneck features to simulate domain shift, yielding stochastic rank maps. Across perturbations, consistently top-graded voxels form stable high-probability ($M_{HP}$) core, while persistently non-background voxels form the stable non-background region ($M_{HR}$). (d) Recursive Self-Evolution: guided by these hints, successive model generations progressively squeeze the ambiguity region between $M_{HP}$ and $M_{HR}$, progressively bringing the two boundaries closer until convergence forms the final segmentation.}
\label{fig:1} 
\end{figure*}

Deep learning has demonstrated state-of-the-art performance in medical image segmentation, crucial for computer-aided diagnosis~\cite{chu2026horuseye,zhou2020rapid,litjens2017survey,ronneberger2015u,jiangtao2025comprehensive}. However, its widespread clinical utility is constrained by the distribution shift: models trained on source databases often suffer from significant performance degradation when deployed to unseen clinical environments, undermining confidence in clinical utility~\cite{lawrence2025artificial,gu2021domain}. This problem further widens the social gap, as models optimized for developed nations fail to accommodate the racial and geographic diversity of developing regions. Practically, as shown in Fig.~\ref{fig:1}(a), distribution shift can be classified as~\cite{yoon2024domain}: 1) Demographic Shift, arising from patient heterogeneity in age, gender, and disease distribution; and 2) Protocol and Modality Shift, driven by acquisition diversity such as variable imaging protocols, noise, and resolution, which collectively hinder cross-site transferability.

To address distribution shift, existing research can be broadly categorized into two paradigms: Training-time Generalization, which focuses on learning invariant representations~\cite{cheng2025dynamic,liu2021feddg,liu2025spectrum,carlucci2019domain}, and Deployment-time Adaptation, which dynamically updates model parameters during inference~\cite{zhang2024iplc,zhou2025tegda,chen2025dual,nado2020evaluating}. However, we argue that these approaches are fundamentally limited by a methodology of exhaustive adaptation. They attempt to either simulate an infinite space of shifts or reactively tune parameters, which struggles in the unpredictable real world. In contrast, our framework is built upon an empirical law, the ``Rank Stability of Positive Regions'', which states that the relative rank of positive predictions remains stable under distribution shifts. As shown in Fig.~\ref{fig:1}(b), although the predicted probability of a lesion may change substantially (for example, from 0.9 to 0.4) under distribution shift, the relative rank ordering of the true positive region remains largely preserved, enabling rank-based segmentation to correctly retain the lesion even under severe distribution shift.

This perspective has recently inspired rudimentary applications in classification tasks. Previous studies have shown that the relative order is the most reliable discriminative indicator to enhance robustness against domain shifts, even under severe model miscalibration in complex environments~\cite{jing2023order,luo2024trustworthy}. However, these works are limited to sample-level class rankings, and the potential of this mechanism in segmentation remains unexplored. Our work provides the first framework to make segmentation based on rank rather than probabilities.

In summary, this paper makes the following key contributions to the field of medical image segmentation. First, we introduce the ``Rank Stability of Positive Regions'' as a critical empirical law under distribution shift, serving as a reliable anchor against open-world distribution shift. Second, we introduce CRISP, a parameter-free framework to transform this principle into an actionable segmentation strategy, which relies on rank rather than probabilities. CRISP iteratively squeezes the uncertainty gap between High-Precision and High-Recall, forming the final segmentation. Third, extensive evaluations on multi-center cardiac MRI and CT-based lung vessel segmentation demonstrate CRISP's superior robustness, significantly outperforming state-of-the-art methods with striking HD95 reductions of up to 0.14 (7.0\% improvement), 1.90 (13.1\% improvement), and 8.39 (38.9\% improvement) pixels across multi-center, demographic, and modality shifts, respectively.
\section{Method}
\textbf{Problem Definition.}
We study source-only domain generalization for volumetric medical image segmentation. Given a labeled source domain $\mathcal{D}_S=\{(X_S,Y_S)\}$ with input volumes $X\in\mathbb{R}^{H\times W\times D}$ and labels $Y\in\{0,1\}^{H\times W\times D}$, our goal is to generalize to unseen domains experiencing distribution shifts. To this end, we propose the CRISP framework (Fig.~\ref{fig:1}). Grounded in the Rank Stability hypothesis, CRISP simulates domain shift via Latent Feature Perturbation to extract domain-invariant spatial hints: a stable high-probability core ($M_{HP}$) and a stable non-background region ($M_{HR}$). These dual hints drive a Recursive Self-Evolution process, progressively squeezing the uncertainty gap for precise, target-free boundary refinement.
\subsection{Recursive Self-Evolution Training Framework}
\label{sec:2.1}
Our objective is to obtain the final segmentation model $M_K$, 
where $k$ denotes the training phase and $M_k$ the model at phase $k$, 
with $M_1,\dots,M_{K-1}$ serving as intermediate stages and only $M_K$ used for inference.

At phase $k$, the training dataset is defined as 
$D_k = D_{k-1} \cup \{ (X, \mathcal{H}^{(k)}, Y) \}$, 
where $\mathcal{H}^{(k)}$ denotes the spatial hint at phase $k$, 
defined as 
$\mathcal{H}^{(k)} = \{ M_{HP}^{(k)}, M_{HR}^{(k)} \}$,
with $M_{HP}^{(k)}$ (stable high-probability core) 
and $M_{HR}^{(k)}$ (stable non-background region),
where $\mathcal{H}^{(k)}$ is generated from $M_{k-1}$.
Newly generated hints are incorporated at each phase to progressively refine structural priors, driving the two boundaries closer until they converge, i.e.,
$\lim_{k \to \infty} M_{HP}^{(k)} = \lim_{k \to \infty} M_{HR}^{(k)} = \hat{Y}$.

The model $M_1$ is first trained on $D_1$ with initialized hints 
($M_{HP}^{(1)}=\mathbf{0}$, $M_{HR}^{(1)}=\mathbf{1}$), 
and for $k \geq 2$, $M_k$ is trained on $D_k$ initialized from $M_{k-1}$. The final model $M_K$ is recursively applied for $K$ refinement iterations, and the prediction at the final iteration is taken as the segmentation.
\subsection{Latent Feature Perturbation for Constraint Generation}
\label{sec:2.2}
We employ Latent Feature Perturbation to simulate domain shifts under the Rank Stability hypothesis, assuming that genuine structures preserve their relative probability ordering across shifts.

Given an input slice $X_{sl}$ and spatial hints $\mathcal{H}^{(k)}$, the encoder $f_{\text{enc}}$ extracts latent features
\begin{equation}
\mathcal{F} = f_{\text{enc}}(X_{sl}, \mathcal{H}^{(k)}),
\end{equation}
representing the deep bottleneck. To simulate domain shift, Gaussian noise $\xi^{(n)} \sim \mathcal{N}(0, \sigma^2 I)$ is injected:
\begin{equation}
\mathcal{F}^{(n)} = \mathcal{F} + \xi^{(n)}, \quad n=1,\dots,N .
\end{equation}
The decoder $f_{\text{dec}}$ maps $\mathcal{F}^{(n)}$ to pixel-wise probabilities
\begin{equation}
P^{(n)} = \mathrm{Softmax}(f_{\text{dec}}(\mathcal{F}^{(n)})),
\end{equation}
where $P^{(n)}_{c}(x,y)$ denotes the class-$c$ probability at pixel $(x,y)$.

To map the predictions into a space that linearly reflects the accumulated amount of evidence rather than bounded probability values, we transform the probabilities to log-odds
\begin{equation}
S_c^{(n)} = \ln \frac{P_c^{(n)}}{1-P_c^{(n)}},
\end{equation}
and discretized into $L$ confidence grades using instance-level statistics:
\begin{equation}
S_{\min}^{(n)} = \min_{c,x,y} S_c^{(n)}(x,y), \quad 
S_{\max}^{(n)} = \max_{c,x,y} S_c^{(n)}(x,y).
\end{equation}
The grade map is defined as
\begin{equation}
G_c^{(n)} =
\left\lfloor
\frac{S_c^{(n)} - S_{\min}^{(n)}}{S_{\max}^{(n)} - S_{\min}^{(n)}} \times L
\right\rfloor_{\text{clamped}},
\quad
G_c^{(n)}(x,y) \in \{0,\dots,L-1\},
\end{equation}
where $\lfloor\cdot\rfloor_{\text{clamped}}$ clips values into $\{0,\dots,L-1\}$. Higher grades indicate stronger foreground confidence.

From the $N$ stochastic grade maps, we derive the class-wise masks $M_{HP,c}^{(k)}$ and $M_{HR,c}^{(k)}$ for each class $c$:
\begin{equation}
M_{HP,c}^{(k)}
=
\bigcap_{n=1}^{N}
\{G_c^{(n)} = L-1\},
\quad
M_{HR,c}^{(k)}
=
\bigcup_{n=1}^{N}
\{G_c^{(n)} > 0\}.
\end{equation}
$M_{HP,c}^{(k)}$ captures the invariant core, while $M_{HR,c}^{(k)}$ defines persistent foreground support. The uncertainty region is
\begin{equation}
\Omega_{u,c}^{(k)} = M_{HR,c}^{(k)} \cap (1 - M_{HP,c}^{(k)}),
\end{equation}
localizing ambiguous boundary pixels for refinement.

\subsection{Accelerated Convergence: Uncertainty Squeezing Loss}
\label{sec:2.3}
To mitigate boundary ambiguity between $M_{HR,c}^{(k)}$ and $M_{HP,c}^{(k)}$ and accelerate refinement, we introduce an Uncertainty Squeezing Loss ($\mathcal{L}_{\text{squeeze}}$) to enforce predictive consistency within the global uncertainty region $\Omega_u^{(k)} = \bigcup_{c=1}^{C} \Omega_{u,c}^{(k)}$:$$\mathcal{L}_{\text{squeeze}} = \frac{1}{|\Omega_u^{(k)}| + \epsilon} \sum_{n=1}^{N} \left\| \left( P_k - P_k^{(n)} \right) \odot \Omega_u^{(k)} \right\|^2,$$where $P_k$ and $P_k^{(n)}$ denote the predictions at generation $k$ and under the $n$-th perturbation, respectively, and $\odot$ is element-wise multiplication. The mask $\Omega_u^{(k)}$ restricts optimization to ambiguous regions, stabilizing boundary refinement.

The overall objective combines standard Dice loss with the proposed squeezing loss:$$\mathcal{L}_{\text{total}} = \mathcal{L}_{\text{dice}} + \alpha \mathcal{L}_{\text{squeeze}},$$where $\mathcal{L}_{\text{dice}}$ denotes the Dice loss between the prediction $P_k$ and the target label, and $\alpha$ is a weighting hyperparameter that balances the contribution of the uncertainty squeezing.

\section{Experiments}
\textbf{Datasets.}
We conducted experiments on one public benchmark and two real-world clinical datasets.
(1) The M\&MS dataset~\cite{campello2021multi} contains cardiac MRI scans from four vendors—Siemens (A), Philips (B), GE (C), and Canon (D)—with 192, 252, 150, and 100 volumes, respectively. Domain A served as the source, while B–D were targets, forming a multi-center shift. Segmentation targets include LV, RV, and MYO. Slices were resized to $256 \times 256$. (2) A CT-based lung vessel dataset includes 93 Contrast-enhanced CT and 101 Non-contrast CT volumes, representing a modality shift due to contrast agent. Contrast-enhanced CT served as the source and Non-contrast CT as the target. Slices were resized to $512 \times 512$. (3) A CT-based lung vessel dataset comprises 98 Normal and 74 COVID-19 cases, reflecting a demographic shift. The Normal Cohort served as the source and the COVID Cohort as the target. The same preprocessing as in (2) was applied. All data collection procedures were approved by the relevant ethics committees, and informed consent was obtained from all participants in accordance with the Declaration of Helsinki.
\begin{table*}[t]
\centering
\caption{Quantitative comparison on M\&Ms dataset. Dice score (\%) $\uparrow$ and HD95 (pixel) $\downarrow$. The best results among domain generalization methods are highlighted in bold. \textcolor{green}{$^{\star}$} = Methods that cannot access target domain. \textcolor{red}{$^{\#}$} = Methods that need target domain to update parameters(need more information, need retraining).}
\label{tab:results}
\renewcommand{\arraystretch}{1.3} 
\resizebox{\textwidth}{!}{
\begin{tabular}{l|l|ccc|ccc|ccc}
\hline
\multirow{2}{*}{Metrics} & \multirow{2}{*}{Method} & \multicolumn{3}{c|}{Target domain B} & \multicolumn{3}{c|}{Target domain C} & \multicolumn{3}{c}{Target domain D} \\ \cline{3-11}
 &  & LV & MYO & RV & LV & MYO & RV & LV & MYO & RV \\ \hline
\multirow{7}{*}{\begin{tabular}[c]{@{}l@{}}Dice $\uparrow$\\ (\%)\end{tabular}} 
 & Baseline & 85.22±13.27 & 75.91±11.22 & 78.25±18.15 & 83.32±11.86 & 74.54±9.15 & 78.30±14.84 & 88.36±8.22 & 75.76±5.44 & 83.57±10.21 \\
 & Upper Bound & 90.92±7.09 & 85.10±3.73 & 86.47±9.39 & 88.60±8.58 & 81.54±7.12 & 85.78±7.12 & 92.75±4.79 & 81.67±3.30 & 87.56±17.83 \\ \cline{2-11}
 & \textcolor{green}{FedDG$^{\star}$}~\cite{liu2021feddg} & 86.97±12.18 & 79.25±5.88 & 85.05±9.72 & 87.17±7.99 & 76.96±7.65 & 83.65±10.20 & 89.91±5.86 & 76.83±4.26 & 86.89±6.71 \\
 & \textcolor{green}{DDG-Med$^{\star}$}~\cite{cheng2025dynamic} & 89.01±9.78 & 80.15±5.08 & 84.81±9.51 & 88.67±6.70 & 78.62±6.39 & 83.33±8.55 & 90.75±3.27 & 78.40±4.94 & 85.34±7.11 \\
 & \textcolor{red}{IPLC$^{\#}$}~\cite{zhang2024iplc} & 88.50±6.23 & \textbf{83.27±3.70} & 83.40±9.01 & 88.62±9.32 & 78.82±7.40 & 84.08±14.39 & 89.91±5.63 & 79.24±3.64 & 86.28±6.99 \\
 & \textcolor{red}{TEGDA$^{\#}$}~\cite{zhou2025tegda} & 88.98±9.28 & 83.21±3.43 & 85.52±7.60 & 85.57±9.64 & 78.01±7.24 & 82.90±10.71 & 89.58±7.10 & 79.58±3.68 & \textbf{87.46±4.92} \\
 & \textcolor{green}{Ours$^{\star}$} & \textbf{89.57±8.55} & 82.74±4.19 & \textbf{86.49±8.53} & \textbf{88.92±6.32} & \textbf{79.82±8.41} & \textbf{84.46±10.03} & \textbf{92.28±3.04} & \textbf{80.81±4.51} & 85.26±7.61 \\ \hline
\multirow{7}{*}{\begin{tabular}[c]{@{}l@{}}HD95 $\downarrow$\\ (pixel)\end{tabular}} 
 & Baseline & 2.64±2.68 & 2.15±1.22 & 3.80±5.88 & 3.57±4.57 & 2.66±1.87 & 3.94±5.90 & 2.25±2.75 & 2.05±0.81 & 2.55±2.87 \\
 & Upper Bound & 1.72±1.47 & 1.55±0.62 & 2.08±1.64 & 2.84±4.04 & 2.39±2.37 & 2.42±1.54 & 2.92±4.77 & 1.45±0.48 & 1.48±1.28 \\ \cline{2-11}
 & \textcolor{green}{FedDG$^{\star}$}~\cite{liu2021feddg} & 2.47±2.03 & 2.17±1.05 & 3.65±5.53 & 2.96±3.05 & 2.32±0.95 & 3.11±2.45 & 2.53±3.44 & 2.12±0.83 & 2.98±5.47 \\
 & \textcolor{green}{DDG-Med$^{\star}$}~\cite{cheng2025dynamic} & 2.43±3.38 & 2.23±1.16 & 3.03±2.83 & 2.29±1.62 & \textbf{2.04±0.80} & 3.05±2.46 & \textbf{1.78±0.59} & 2.30±0.92 & 3.65±3.54 \\
 & \textcolor{red}{IPLC$^{\#}$}~\cite{zhang2024iplc} & \textbf{1.74±1.25} & 1.86±1.17 & 2.05±1.46 & 2.97±3.40 & 2.17±1.39 & 2.92±3.15 & 1.79±2.01 & 1.80±0.52 & 2.04±1.82 \\
 & \textcolor{red}{TEGDA$^{\#}$}~\cite{zhou2025tegda} & 1.88±1.99 & \textbf{1.53±0.56} & 1.86±1.17 & 2.65±2.43 & 2.49±1.74 & 2.40±2.28 & 2.10±3.20 & 1.66±0.45 & 1.44±0.52 \\
 & \textcolor{green}{Ours$^{\star}$} & 1.80±1.30 & 1.57±0.64 & \textbf{1.57±1.21} & \textbf{2.06±1.63} & 2.17±1.60 & \textbf{2.20±3.05} & 2.43±4.57 & \textbf{1.60±0.53} & \textbf{1.38±0.86} \\ \hline
\end{tabular}
}
\end{table*}

\noindent\textbf{Implementation Details.} Following previous works~\cite{liu2025spectrum,chen2025dual}, we employ DeepLa\-bv3+~\cite{chen2018encoder} with a MobileNetV2~\cite{sandler2018mobilenetv2} backbone to ensure a fair comparison. All experiments are implemented in PyTorch on an A6000 GPU and optimized for 60,000 iterations using Adam (learning rate 0.001), with a batch size of 10 for M\&MS~\cite{campello2021multi} and 35 for private CT datasets. For CRISP, we set the latent perturbation steps to $N=10$, noise standard deviation to $\sigma=0.35$, confidence grades to $L=5$, and the squeezing loss weight to $\alpha=0.5$. Performance is evaluated via volume-level Dice and HD95.

\noindent\textbf{Comparison with SOTAs.} Our CRISP method was compared with four state-of-the-art methods on the M\&MS dataset: two leading domain generalization (DG) models, including FedDG~\cite{liu2021feddg} and DDG-Med~\cite{cheng2025dynamic}, and two deployment-time adaptation strategies, including the SFDA method IPLC~\cite{zhang2024iplc} and the TTA approach TEGDA~\cite{zhou2025tegda}. We also evaluate against two baseline scenarios: 1) Baseline, where the pre-trained source model is directly tested on target domains; and 2) Upper Bound, which provides a performance ceiling by jointly training on source and target data (e.g., $A+B$ for domain $B$). Notably, both the Upper Bound and IPLC are evaluated via 5-fold cross-validation.

The quantitative evaluation results are shown in Table~\ref{tab:results}. Taking target domain C as an example, compared with the Source baseline, our method achieved 
\begin{table*}[t]
\centering
\caption{Quantitative evaluation across different clinical scenarios. \textbf{(S)}: Source domain, \textbf{(T)}: Target domain. Dice score (\%) $\uparrow$ and HD95 (pixel) $\downarrow$. \textcolor{green}{$^{\star}$} = Methods that cannot access target domain. \textcolor{red}{$^{\#}$} = Methods that need target domain to update parameters(need more information, need retraining).}
\label{tab:combined_scenarios}
\renewcommand{\arraystretch}{1} 
\resizebox{1.0\textwidth}{!}{
\begin{tabular}{l|l|cc|cc}
\hline
\multirow{2}{*}{Metrics} & \multirow{2}{*}{Method} & \multicolumn{2}{c|}{Scenario 1 (Demographic)} & \multicolumn{2}{c}{Scenario 2 (Modality)} \\ \cline{3-6} 
 &  & Normal Cohort (S) & COVID Cohort (T) & Contrast-enhanced (S) & Non-contrast (T) \\ \hline
\multirow{4}{*}{\begin{tabular}[c]{@{}l@{}}Dice $\uparrow$\\ (\%)\end{tabular}} 
 & Baseline & 81.14±4.24 & 65.15±16.62 & 74.47±3.86 & 33.26±8.48 \\
 & \textcolor{red}{IPLC$^{\#}$}~\cite{zhang2024iplc} & 81.14±4.24 & 59.92±6.56 & 74.47±3.86 & 41.33±11.27 \\
 & \textcolor{red}{TEGDA$^{\#}$}~\cite{zhou2025tegda} & 81.14±4.24 & 66.65±11.46 & 74.47±3.86 & 36.30±6.99 \\
 & \textcolor{green}{Ours$^{\star}$} & \textbf{83.73±5.47} & \textbf{70.61±13.27} & \textbf{78.23±2.94} & \textbf{59.46±7.03} \\ \hline
\multirow{4}{*}{\begin{tabular}[c]{@{}l@{}}HD95 $\downarrow$\\ (pixel)\end{tabular}} 
 & Baseline & 4.84±2.43 & 17.12±25.04 & 8.24±2.78 & 29.97±13.38 \\
 & \textcolor{red}{IPLC$^{\#}$}~\cite{zhang2024iplc} & 4.84±2.43 & 19.35±28.57 & 8.24±2.78 & 21.58±11.86 \\
 & \textcolor{red}{TEGDA$^{\#}$}~\cite{zhou2025tegda} & 4.84±2.43 & 14.47±22.08 & 8.24±2.78 & 24.39±13.72 \\
 & \textcolor{green}{Ours$^{\star}$} & \textbf{2.49±1.99} & \textbf{12.57±23.31} & \textbf{6.76±1.83} & \textbf{13.19±11.07} \\ \hline
\end{tabular}
}
\end{table*}
Dice scores of 88.92\%, 79.82\%, and 84.46\% for the LV, MYO, and RV, respectively, representing performance improvements of 5.60\%, 5.28\%, and 6.16\%. Notably, CRISP even surpasses the Upper Bound across the majority of categories in terms of boundary precision, achieving an overall average HD95 reduction of 0.23 pixels across all target domains compared to the fully-supervised ceiling.

\begin{figure}[H] 
\centering 
\includegraphics[width=1.0\textwidth]{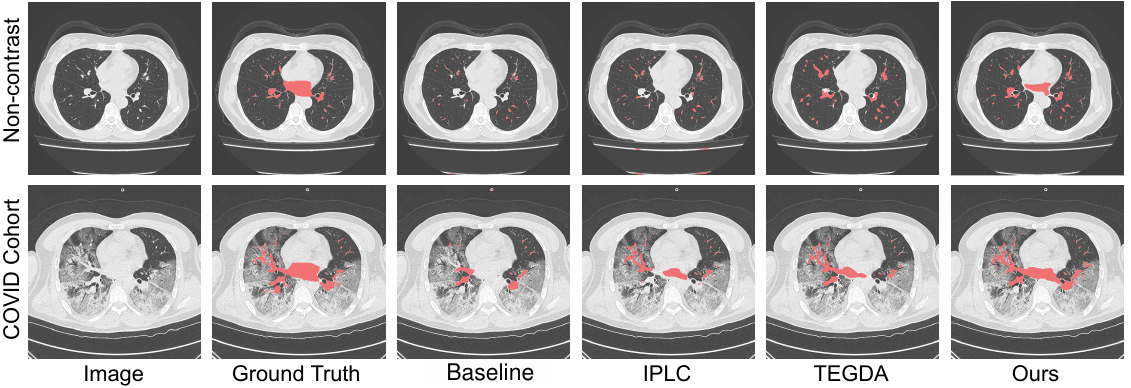} 
\caption{Qualitative segmentation results of different methods. The top and bottom rows present the segmentation performance under modality shift (Non-contrast) and demographic shift (COVID Cohort), respectively.} 
\label{fig:2} 
\end{figure}

As shown in Table~\ref{tab:combined_scenarios}, we further evaluate the generalization capability of CRISP under two scenarios involving severe shifts: Demographic shift (Normal Cohort $\rightarrow$ COVID Cohort) and modality shifts (Contrast-enhanced $\rightarrow$ Non-contrast). Given the superior performance of adaptation-based methods, we focus our comparison on their performance in these two challenging target domains. Quantitatively, CRISP surpasses the best competing adaptation method by 3.96\% and 18.13\% in Dice score, and by 1.90 and 8.39 in HD95 for the COVID Cohort and Non-contrast target domains, respectively. A visual comparison between different adaptation-based methods is shown in Fig.~\ref{fig:2}. These results indicate that by leveraging inherent anatomical stability rather than chasing distribution shifts, CRISP achieves high-precision boundary refinement without requiring any target domain information.

\begin{figure}[t]
    \centering
    \begin{minipage}[t]{0.48\textwidth} 
        \vspace{0pt} 
        \centering
        \begin{minipage}[t]{0.48\linewidth}
            \vspace{0pt}
            \centering
            \includegraphics[width=\linewidth]{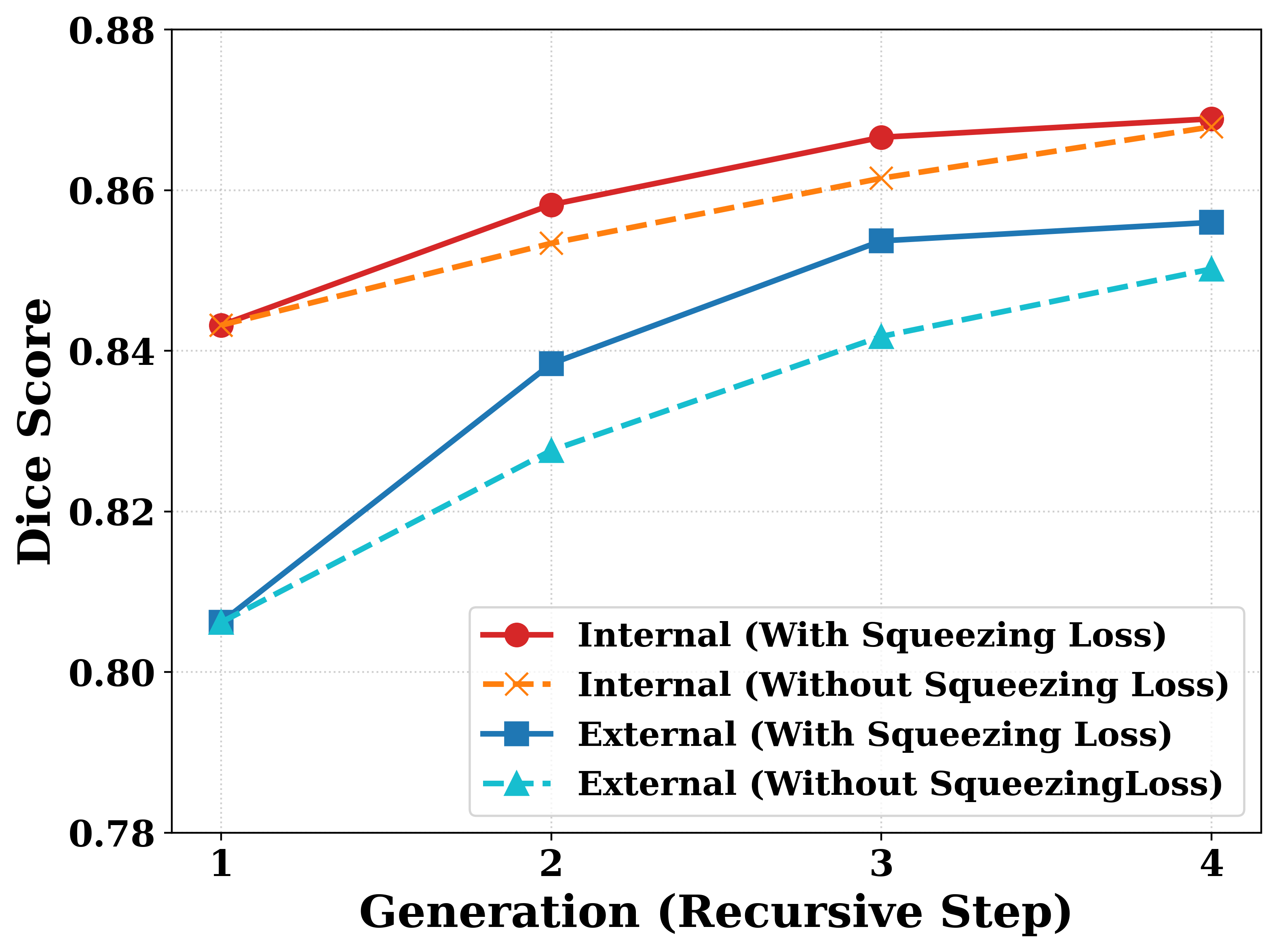}
        \end{minipage}
        \hfill
        \begin{minipage}[t]{0.48\linewidth}
            \vspace{0pt}
            \centering
            \includegraphics[width=\linewidth]{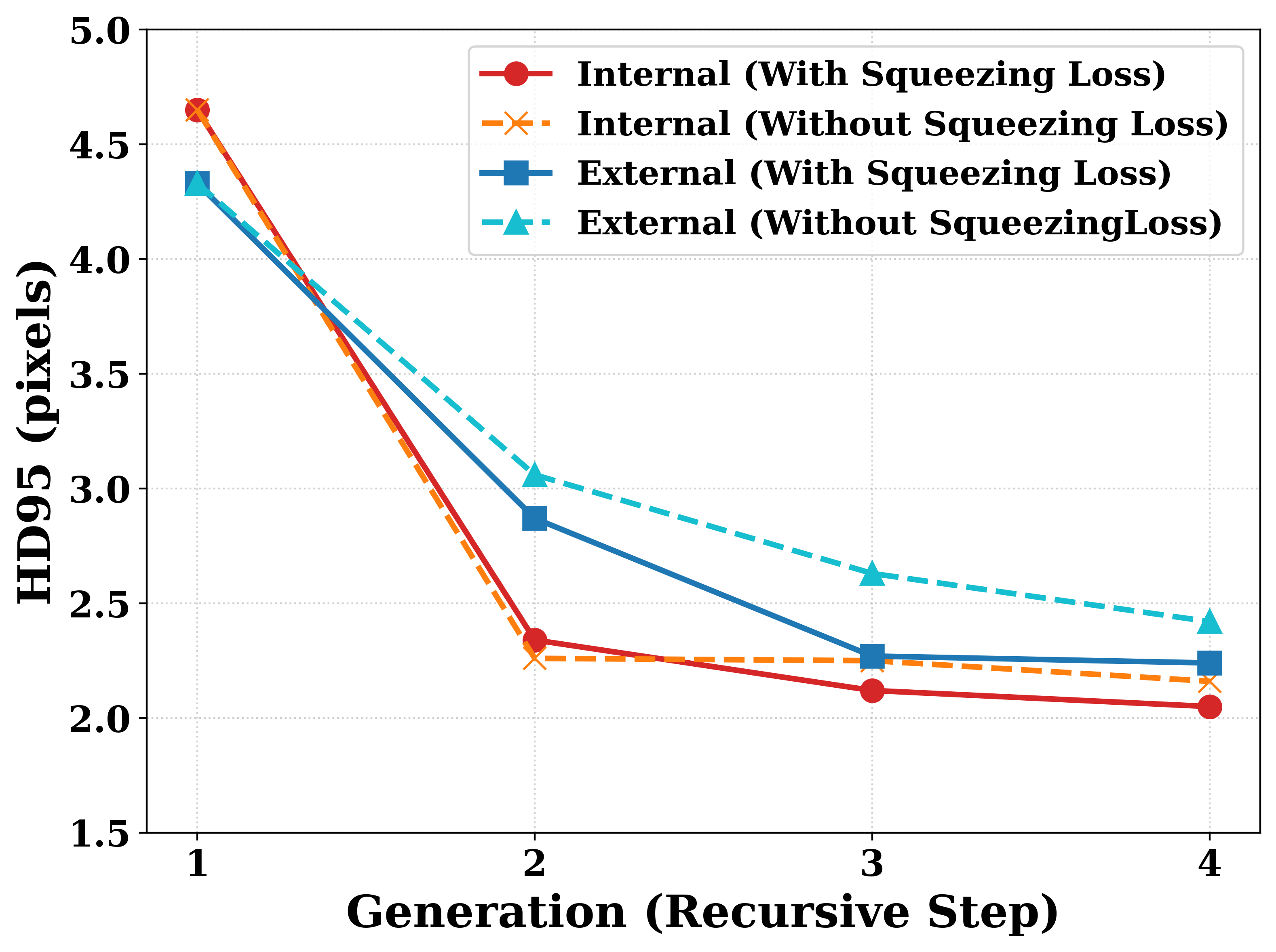}
        \end{minipage}
        
        \captionof{figure}{Convergence analysis of Dice (left) and HD95 (right) scores.}
        \label{fig:ablation1}
    \end{minipage}
    \hfill 
    \begin{minipage}[t]{0.48\textwidth}
        \vspace{0pt} 
        \centering
        \captionof{table}{Ablation study of CRISP on three distinct challenges.}
        \label{tab:ablation}
        
        \setlength{\tabcolsep}{1.5pt} 
        \resizebox{\linewidth}{!}{ 
        \begin{tabular}{cccccccc}
        \toprule
        \multicolumn{1}{c}{Recursive} &
        \multicolumn{1}{c}{Uncertainty} &
        \multicolumn{2}{c}{M\&MS} &
        \multicolumn{2}{c}{Non-contrast} &
        \multicolumn{2}{c}{COVID Cohort} \\
        \multicolumn{1}{c}{Self-Evolution} &
        \multicolumn{1}{c}{Squeezing Loss} &
        Dice$\uparrow$ & HD95$\downarrow$ &
        Dice$\uparrow$ & HD95$\downarrow$ &
        Dice$\uparrow$ & HD95$\downarrow$ \\
        \midrule
         &  &
         \makecell{80.62 \\ $\pm$2.54} & \makecell{4.33 \\ $\pm$1.62} &
         \makecell{33.26 \\ $\pm$8.48} & \makecell{29.97 \\ $\pm$13.38} &
         \makecell{65.15 \\ $\pm$16.62} & \makecell{17.12 \\ $\pm$25.04} \\
        \ding{51} &  &
         \makecell{85.02 \\ $\pm$0.56} & \makecell{2.42 \\ $\pm$0.13} &
         \makecell{51.45 \\ $\pm$6.57} & \makecell{19.80 \\ $\pm$12.86} &
         \makecell{67.42 \\ $\pm$12.48} & \makecell{16.27 \\ $\pm$23.53} \\
        \ding{51} & \ding{51} &
        \makecell{\textbf{85.60} \\ \textbf{$\pm$0.85}} &
        \makecell{\textbf{2.24} \\ \textbf{$\pm$0.42}} &
        \makecell{\textbf{59.46} \\ \textbf{$\pm$7.03}} &
        \makecell{\textbf{13.19} \\ \textbf{$\pm$11.07}} &
        \makecell{\textbf{70.61} \\ \textbf{$\pm$13.27}} &
        \makecell{\textbf{12.57} \\ \textbf{$\pm$23.31}} \\

        \bottomrule
        \end{tabular}

        }
    \end{minipage}
\end{figure}

\textbf{Ablation Study.} 
As shown in Fig.~\ref{fig:ablation1}, we investigated the convergence dynamics of CRISP within the M\&MS target domain. While standard self-adaptation processes typically require $N$ generations to stabilize, our proposed method---incorporating Uncertainty Squeezing Loss---achieves significant early convergence by the 3rd generation, indicating accelerated optimization.

To verify the effectiveness of our framework, we conducted ablation experiments in Table~\ref{tab:ablation}, comparing three configurations: (1) a baseline without adaptation; (2) an iterative adaptation driven by Recursive Self-Evolution; and (3) the complete CRISP strategy. The baseline showed clear vulnerability under multi-center, modality, and demographic shifts. Introducing Recursive Self-Evolution significantly improved generalization across these distribution shifts. Moreover, the full CRISP framework achieved the highest gains on all datasets, with particularly pronounced improvements under modality and demographic shifts. These results confirm that the Uncertainty Squeezing Loss effectively refines boundary priors and facilitates recursive adaptation to complex target tasks.

\section{Conclusion}
We proposed CRISP, a parameter-free and model-agnostic framework for medical image segmentation, addressing generalization bottlenecks under severe distribution shifts. Unlike current paradigms that attempt to exhaust infinite real-world shifts, CRISP leverages a fundamental empirical law—the Rank Stability of Positive Regions—to anchor anatomy in source-only settings, performing segmentation based on relative rank rather than absolute probabilities. Specifically, Latent Feature Perturbation simulates model behavior under shift to construct High-Precision and High-Recall priors, while a recursive squeezing dynamic, guided by the Uncertainty-Squeezing Loss, refines boundaries and accelerates convergence. Experiments on cardiac MRI and CT-based lung vessel segmentation under protocol, modality, and demographic shifts demonstrate its superior robustness over state-of-the-art methods. Future work will explore unrolling the recursive process into a unified, stochastic training loop. Training on mixed-stage batches enables the model to learn a universal squeezing policy for efficient self-convergence.

%
%
%
%

\end{document}